\documentclass{article}
\usepackage{spconf,amsmath,graphicx}

\usepackage{amsmath,graphicx}
\usepackage{amssymb,amsmath,epsfig}
\usepackage{lipsum,color,multirow,url}
\usepackage{arydshln, algorithm, algpseudocode}
\usepackage{microtype}
\usepackage{nth}
\usepackage[inline]{enumitem}
\usepackage{romannum}
\usepackage{textcomp}

\title{Improving RNN Transducer Based ASR with Auxiliary Tasks}
\name{Chunxi Liu,   Frank Zhang, Duc Le, Suyoun Kim,  Yatharth Saraf, Geoffrey Zweig}     \address{Facebook AI, USA \\
{\small \tt \{chunxiliu,frankz,duchoangle,suyounkim,ysaraf,gzweig\}@fb.com}  }

\begin{document}
\ninept
\maketitle
\begin{abstract}
End-to-end automatic speech recognition (ASR) models with a single neural network have recently demonstrated state-of-the-art results compared to conventional hybrid speech recognizers. Specifically, recurrent neural network transducer (RNN-T) has shown competitive ASR performance on various benchmarks.
In this work, we examine ways in which RNN-T can achieve better ASR accuracy via performing auxiliary tasks.
We propose (i) using the same auxiliary task as primary RNN-T ASR task, and (ii) performing context-dependent graphemic state prediction as in conventional hybrid modeling.
In transcribing social media videos with varying training data size, we first evaluate the streaming ASR performance on three languages: Romanian, Turkish and German. We find that both proposed methods provide consistent improvements. 
Next, we observe that both auxiliary tasks demonstrate efficacy in learning deep transformer encoders for RNN-T criterion, thus achieving competitive results -  2.0\%/4.2\% WER on LibriSpeech test-clean/other - as compared to prior top performing models.
 % e.g. on-device applications and long-form speech recognition  tasks and 
\end{abstract}
%
% ------------------------------------------------------------------------------------------------------------------------------------------------
\section{Introduction}
\label{sec:intro}

Building conventional hidden Markov model (HMM) based hybrid automatic speech recognition (ASR) systems include multiple engineered steps like bootstrapping, decision tree clustering of context-dependent phonetic/graphemic states \cite{young1994tree}, acoustic and language model training, etc. 
End-to-end ASR models \cite{graves2012sequence, chan2016listen, zweig2017advances, sak2017recurrent} use neural networks to transduce audio into word sequences, and can be learned in a single step from scratch.
Specifically, recurrent neural network transducer (RNN-T) originally presented in \cite{graves2012sequence} – also referred to as sequence transducer – has been shown preferable on numerous applications. % , especially when streaming ASR is required.
For example, the model size of RNN-T is much more compact than conventional hybrid models, being favorable as an on-device recognizer \cite{he2019streaming, li2019improving, sainath2020streaming}. It also has been demonstrated as a high-performing streaming model in extensive benchmarks \cite{battenberg2017exploring, chiu2019comparison, li2020comparison, zhang2021benchmarking}. 
Such recent success has motivated the efforts to improve RNN-T from various aspects, e.g. model pretraining \cite{rao2017exploring, hu2020exploring}, generalization ability on long-form audios \cite{narayanan2019recognizing}, training algorithms \cite{li2019improving, weng2019minimum}, speech enhancement \cite{pandey2021dual}, etc. 

In this work, we make an attempt on improving RNN-T via auxiliary learning, which aims to improve the generalization ability of a primary task by training on additional auxiliary tasks \cite{jaderberg2016reinforcement, liu2019self}. 
While multitask learning \cite{caruana1997multitask} may aim to improve the performance of multiple tasks simultaneously, auxiliary learning selectively serves to assist the primary task and only the primary task performance is in focus.  
Auxiliary learning has been studied extensively in reinforcement learning \cite{jaderberg2016reinforcement, du2018adapting}, where pseudo-reward functions are designed to enable the main policy to be learned more efficiently.      
In the context of attention-based sequence-to-sequence (seq2seq) ASR models, \cite{toshniwal2017multitask, moriya2018multi} show that 
learning encoders with auxiliary tasks of predicting phonemes or context-dependent phonetic HMM states (i.e. senones \cite{dahl2012context}) can improve the primary ASR word error rate (WER).  
\cite{zaremoodi2019adaptively} shows that using auxiliary syntactic and semantic tasks can improve the main low-resource machine translation task.

In this paper, we consider the application of auxiliary tasks to RNN-T based ASR. First, we design an auxiliary task to be the same ASR task, where the transducer encoder forks from an intermediate encoder layer, and both the primary branch and auxiliary branch perform ASR tasks.
Note that in this way, both primary and auxiliary branches can provide posterior distributions over output labels – characters or wordpieces. Inspired by the prior works \cite{zhang2018deep, lu2019self, li2020dynamic}, we exploit a symmetric Kullback–Leibler (KL) divergence loss between the output posterior distributions of primary and auxiliary branches, along with the standard RNN-T loss. 
Such mutual KL divergence loss is expected to implicitly penalize the inconsistent gradients from the primary and auxiliary losses with respect to their shared parameters, and relieve the optimization inconsistency across tasks \cite{li2020dynamic}. Overall, the knowledge distilled from auxiliary tasks help a model learn better representations shared between primary and auxiliary branches, by enabling the model to find a more robust (flatter) minima and to better generalize to test data \cite{zhang2018deep}.

Secondly, we propose an alternative auxiliary task of predicting context-dependent graphemic states, also referred to as chenones \cite{le2019senones}, as in standard HMM-based hybrid modeling. Similar to the auxiliary senone classification for improving attention-based seq2seq model \cite{toshniwal2017multitask, moriya2018multi}, we exploit chenone prediction for improving RNN-T without relying on language-specific phonemic lexicon. 
HMM-based graphemic hybrid ASR systems have been shown to achieve comparable performance to phonetic lexicon based approaches \cite{kanthak2002context, gales2015unicode, le2019senones}, and still demonstrate state-of-the-art results on common benchmarks when compared to end-to-end models \cite{wang2019transformer}. 
In this paper, we examine if the context-dependent graphemic knowledge – from a decision tree clustering of tri-grapheme HMM states – can be complementary to the character or wordpiece (i.e. subword unit) modeling used in end-to-end ASR \cite{rao2017exploring}, and if the auxiliary chenone prediction task provides an avenue of distilling such context-dependent graphemic knowledge into RNN-T training by providing additional discriminative information. 

To evaluate our proposed methods, we first use streamable ASR models on a challenging task of transcribing social media videos, in both low-resource 
(training data size {\fontfamily{pcr}\selectfont\texttildelow}160 hours) and medium-resource ({\fontfamily{pcr}\selectfont\texttildelow}3K hours) conditions. 
% We observe 3 - 6\% relative improvement by using auxiliary RNN-T loss and symmetric KL divergence in combination, and 3 - 8\% by doing auxiliary chenone prediction. 
Next, on LibriSpeech, we consider the application of auxiliary tasks to the sequence transducers built with deep transformer encoders. 

%
%   {\fontfamily{ptm}\selectfont\texttildelow}160        \texttildelow160
% state-of-the-art results on LibriSpeech \cite{zhang2020transformer} and  production data      CTC \cite{graves2006connectionist}
% ------------------------------------------------------------------------------------------------------------------------------------------------
% ------------------------------------------------------------------------------------------------------------------------------------------------
\section{Modeling Approaches}
\label{sec:model}

In this section we begin with a review of RNN-T based ASR, as originally presented in \cite{graves2012sequence}. Then we present our proposed auxiliary RNN-T task. 
Lastly, we describe the auxiliary context-dependent graphemic state prediction task.     

% ------------------------------------------------------------------------------------------------------------------------------------------------
\subsection{RNN-T}
\label{ssec:rnnt}

ASR can be formulated as a sequence-to-sequence problem. 
Each speech utterance is parameterized as an input acoustic feature vector sequence
$\textbf{x}  = \{\textbf{x}_1 \ldots \textbf{x}_T\} = \textbf{x}_{1:T} $,  % \{\textbf{x}_t\}_{t=1}^T  
where $\textbf{x}_t \in \mathbb{R}^{d}$ and $T$ is the number of frames.  
Denote a grapheme set or a wordpiece inventory as $\mathcal{Y}$, and
the corresponding output sequence of length $U$ as $\textbf{y} = \{y_1 \ldots y_U\} = \textbf{y}_{1:U} $, where $y_u \in \mathcal{Y}$. 

We define $\bar{\mathcal{Y}}$ as $ \mathcal{Y} \cup \{ \emptyset \}$, where $\emptyset$ is the blank label.
Denote $\bar{\mathcal{Y}}^{*}$ as the set of all sequences over output space $\bar{\mathcal{Y}}$, and the element $\textbf{a} \in  \bar{\mathcal{Y}}^*$ as an alignment sequence.
% of length $\bar{U}$ represents the alignment between each time $t$ and an output label  $y_{\bar{U}}$ in $\bar{\mathcal{Y}}$.   factorize
Then we have the posterior probability as:
\begin{equation}
    P( \textbf{y}  | \textbf{x}) =  \sum\limits_{ \textbf{a}  \in \mathcal{B}^{-1}(\textbf{y} ) }    P( \textbf{a}  | \textbf{x})
\label{eq:posterior}
\end{equation}
\noindent where $\mathcal{B}: \bar{\mathcal{Y}}^* \rightarrow  \mathcal{Y}^{*}  $ is a function that removes blank symbols from an alignment \textbf{a}.
RNN-T model parameterizes the alignment probability $P(\textbf{a} | \textbf{x})$ and computes it with an encoder network (i.e. transcription network in \cite{graves2012sequence}), a prediction network and a joint network. 
The encoder performs a mapping operation, denoted as $f^{\text{enc}}$,  which converts $\textbf{x}$ into another sequence of representations 
$\textbf{h}^{\text{enc}}_{1:T} = \{\textbf{h}_1^{\text{enc}} \ldots \textbf{h}^{\text{enc}}_{T}\}$: 
\begin{equation}
 \textbf{h}^{\text{enc}}_{1:T} = f^{\text{enc}}(\textbf{x}) 
%\label{eq:encoder}
\end{equation}
\noindent
%  where $T'$ is equal or shorter than $T$ due to subsampled frame rate. 
A  prediction network $f^{\text{pred}}$, based on RNN or its variants, takes both its state vector and the previous non-blank output label $y_{u-1}$ predicted by the model, to produce the new representation $\textbf{h}^{\text{pred}}_u$: 
\begin{equation}
    \textbf{h}^{\text{pred}}_{1:u} = f^{\text{pred}}(y_{0:(u-1)})
\end{equation}
\noindent where $u$ is output label index and $y_0 = \emptyset$.
The joint network $f^{\text{join}}$  is a feed-forward network that combines encoder output $\textbf{h}^{\text{enc}}_t$ and prediction network output $\textbf{h}^{\text{pred}}_u$ to compute logits $\textbf{z}_{t,u}$:
\begin{equation}
    \textbf{z}_{t,u} = f^{\text{join}}(\textbf{h}^{\text{enc}}_t, \textbf{h}^{\text{pred}}_u) 
\end{equation}
\begin{equation}
\begin{split} 
    P(y_u|  \textbf{x}_{1:t}  ,  y_{1:(u-1)} )  = \text{Softmax}(\textbf{z}_{t,u})   % \textbf{x}_1 \ldots \textbf{x}_t    y_1 \ldots y_{u-1} 
% & = P(y_u| \textbf{h}^{\text{enc}}_t, \textbf{h}^{\text{pre}}_u   ) \\            
\end{split}
\label{eq:posterior_1}
\end{equation}
\noindent such that the logits go through a softmax function and produce a posterior\footnote{
Note that, the posterior distribution in Eq. \ref{eq:posterior_1} can also be written as 
$ P(y_u| \textbf{x}_{1:T}, y_{1:(u-1)}  )$, if the encoder uses global/infinite context, like a BLSTM or non-streaming transformer network \cite{wang2019transformer, zhang2020transformer}.
}
distribution of the next output label over $\bar{\mathcal{Y}}$. 
Finally, the RNN-T loss function is then the negative log posterior as in Eq. \ref{eq:posterior}:
\begin{equation}
    \mathcal{L}^{\text{\tiny{RNN-T}}} (\theta) = - \log  P(\textbf{y} |  \textbf{x} ) 
\end{equation}
\noindent where $\theta$ denotes the model parameters.  Note that the encoder is analogous to an acoustic model, and the combination of prediction network and joint network can be seen as a decoder.

% ------------------------------------------------------------------------------------------------------------------------------------------------
\subsection{Auxiliary sequence transducer modeling}
\label{ssec:aux}

%During RNN-T training

The RNN-T decoder can be viewed as a RNN language model. The RNN takes both its state vector and $y_{u-1}$ to predict $y_u$, so implicitly predicting $y_u$ is conditioned on the whole label history $y_1 \ldots y_{u-1}$ as in  Eq. \ref{eq:posterior_1}. Since the label history can be very informative in predicting the next output label, we conjecture that the posterior entropy over $\bar{\mathcal{Y}}$ computed by Eq. \ref{eq:posterior_1} may be excessively reduced, resulting in encoder undertraining. In other words, if the decoder has played a major role in predicting each $y_u$ by such teacher forcing procedure, which can still result in a reasonable training loss, the encoder may underfit the input $\textbf{x}$,
% not capture the underlying structure of , 
and the resulting generalization can be worse than a model with an adequately trained encoder.  

Additionally, gradient flow \cite{hochreiter2001gradient} through a deep neural network architecture is difficult in general, due to the gradient vanishing/exploding problem at lower layers. 
Although we could add shortcut connections \cite{srivastava2015training, he2016deep} across encoder layers that would help gradient flow through the encoder, 
it does not address the encoder undertraining problem – if the posterior of Eq. \ref{eq:posterior_1} has been peaked at the true label due to the strong cue from previous label history.

\subsubsection{Auxiliary RNN-T criterion}
\label{ssec:aux_loss}

An alternative proposal to increase the gradient signal is based on connecting auxiliary classifiers to intermediate layers directly \cite{szegedy2015going}. 
In this work, to address encoder underfitting and provide the encoder with more backward gradients, we take the approach of connecting an auxiliary branch to an intermediate encoder layer and applying the same RNN-T loss function.

% propagated
% ------------------------------------------------------------------------------------------------------------------------------------------------
\begin{figure}[t]
\centering
\includegraphics[width=0.99\linewidth]{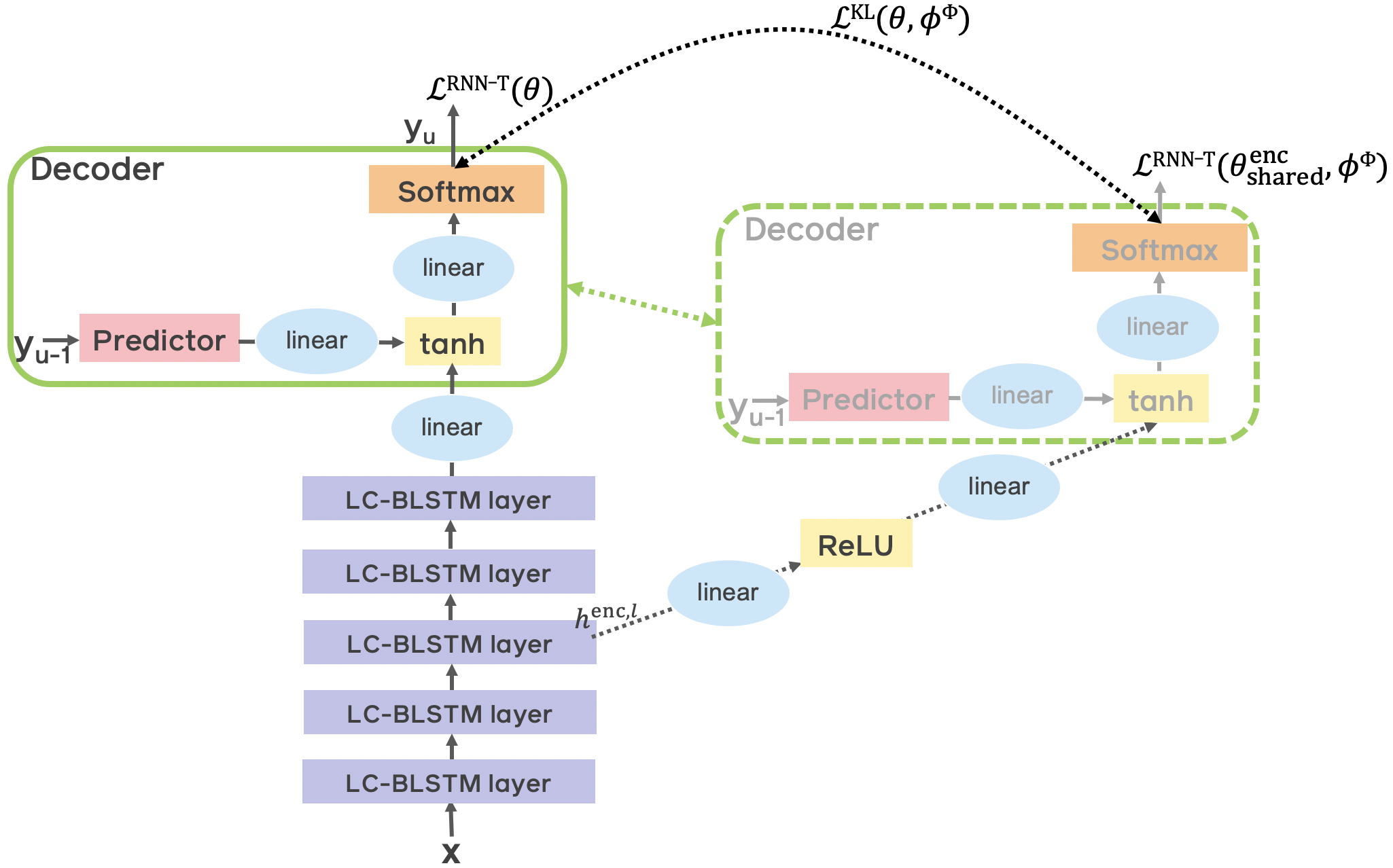}
\caption{\it Illustration of the proposed auxiliary RNN-T and KL divergence criteria. 
For the auxiliary criteria, decoder is shown in a dashed box when it is used by the auxiliary branch to compute the logits (Eq. \ref{eq:mlp}) in the forward pass,
while  the decoder is not updated in the backward pass.   }  %  In other words,
\label{fig:aux}
\end{figure}
% ------------------------------------------------------------------------------------------------------------------------------------------------

As in Figure \ref{fig:aux},
given an $L$-layer encoder network, denote %  $\textbf{h}^{\text{enc}, L}$ as the hidden activations of the top layer, and
$\textbf{h}^{\text{enc}, l}$ as the hidden activations of an intermediate layer $l$, where $1 \leqslant l < L$. 
 $\textbf{h}^{\text{enc}, l}$ goes through a one-hidden-layer multi-layer perceptron (MLP), parameterized by $\phi^l$, and use the same decoder to compute the logits of auxiliary branch: 
% \begin{equation}   \textbf{h}^{\text{enc}, l, \text{aux}}_t  = \text{MLP}(\textbf{h}^{\text{enc}, l}_t )    \label{eq:mlp}   \end{equation}
\begin{equation}
    \textbf{z}_{t,u}^{\text{aux}, l} = f^{\text{join}}(\text{MLP}(\textbf{h}^{\text{enc}, l}_t ), \textbf{h}^{\text{pred}}_u)  
\label{eq:mlp} 
\end{equation}
\begin{equation}
\begin{split}
  P_{\text{aux}, l}(y_u| \textbf{x}_{1:t}  ,  y_{1:(u-1)}  ) 
            % & = P_{\text{aux}}(y_u| \textbf{h}^{\text{enc}, l}_t, \textbf{h}^{\text{pre}}_u   ) \\
            & =  \text{Softmax}(\textbf{z}_{t,u}^{\text{aux}, l}) 
\end{split}
\label{eq:posterior_2}
\end{equation}
\noindent such that we can apply another RNN-T objective function to this auxiliary branch,
and the overall objective function becomes: 
\begin{equation}
\begin{split}
   \mathcal{L} (\theta, \phi) & =  \mathcal{L}^{\text{\tiny{RNN-T}}}  (\theta) +
   \lambda_{\text{aux}}  \mathcal{L}^{\text{\tiny{RNN-T}}} ( \theta_{\text{shared}}^{\text{enc}}, \phi )  \\
  &   = - \log  P(\textbf{y} |  \textbf{x} ) - \lambda_{\text{aux}} \log  P_{\text{aux}, l}(\textbf{y} |  \textbf{x} ) 
\end{split}
\label{eq:objective1}
\end{equation}
\noindent where $\theta$ denotes the parameters of primary branch including the whole encoder and decoder, and $\theta_{\text{shared}}^{\text{enc}}$ denotes the encoder layers $1$ - $l$ shared by primary and auxiliary branches, and $\lambda_{\text{aux}}$ is a weighting parameter.
% The coefficient  scales the auxiliary loss
Note that the auxiliary branch requires a decoder to compute $\textbf{h}^{\text{pred}}_u$ and then $\textbf{z}_{t,u}^{\text{aux}, l}$. Instead of adding another decoder specifically for auxiliary branch, we propose to share the primary decoder during the forward pass; however, we do not update the decoder parameters if the gradients are back propagated from the auxiliary RNN-T loss. Because the auxiliary loss is to address the encoder underfitting issue, decoder is not explicitly learned to fit the auxiliary objective function.

Note that for the auxiliary model, we connect a nonlinear MLP (Eq. \ref{eq:mlp}) – rather than a single linear layer – to the intermediate encoder layer $l$.
Since the lower encoder layers are focused on feature extraction rather than the meaningful final label prediction, directly encouraging discrimination in the low-level representations is suboptimal. 
This is similar to the primary branch, where additional encoder layers of $l+1$ to $L$ are added on top of layer $l$; 
thus, adding the MLP allows for a similar coarse-to-fine architecture, and the shared encoder layers play a more consistent role for both branches.

Finally, when an encoder has a large network depth, we can apply such criterion to multiple encoder layers. 
Denote $\Phi$ as a set of encoder layer indices that are connected with each auxiliary criterion, and $\Phi \subseteq  \{ 1 \ldots L - 1 \}$. 
Denote $I$ as a binary indicator function as 

\begin{equation}
%\quad  = \left\{ 
    I(l) = \left\{
                \begin{array}{ll}
                  1,  \quad    l \in    \Phi    \\   % \hspace{1.82cm}  
                  0,  \quad    l \notin \Phi     \\
                \end{array}
              \right.
\end{equation}
\noindent where  $1 \leqslant l < L$. Then Eq. \ref{eq:objective1} becomes  
\begin{equation}
\begin{split}
   \mathcal{L} (\theta, \phi^{\Phi }) & =  \mathcal{L}^{\text{\tiny{RNN-T}}}  (\theta) +
   \lambda_{\text{aux}}  \mathcal{L}^{\text{\tiny{RNN-T}}} ( \theta_{\text{shared}}^{\text{enc}}, \phi^{\Phi } )  \\
  &   = - \log  P(\textbf{y} |  \textbf{x} ) - \lambda_{\text{aux}} \sum_{l = 1}^{L - 1}   I(l)  \log  P_{\text{aux}, l}(\textbf{y} |  \textbf{x} ) 
\end{split}
\end{equation}

% ------------------------------------------------------------------------------------------------------------------------------------------------
\subsubsection{Auxiliary symmetric KL divergence criterion}
\label{ssec:kl_loss}

Further, prior works \cite{zhang2018deep, lu2019self, li2020dynamic} show that aligning the pairwise posterior distributions of multiple (sub)networks in a mutual learning strategy achieves better performance than learning independently. Thus, other than the supervised learning objective function, i.e. RNN-T criterion, we also explore an additional symmetric KL divergence criterion between the output posterior distributions of both branches: 
\begin{equation}
\begin{split}
   & \mathcal{L}^{\text{KL}} (\theta, \phi^{\Phi } )   =  \frac{ 1 }{ T } \sum\limits_{t=1}^T    \sum_{l = 1}^{L - 1}       \frac{ 1 }{ U } \sum\limits_{u=1}^U    I(l)  \big[   \\
   & D_{\mathrm{KL}}(  P(y_u| \textbf{x}_{1:t}  ,  y_{1:(u-1)}  )  \|  P_{\text{aux}, l}(y_u| \textbf{x}_{1:t}  ,  y_{1:(u-1)}  )  \\
  & +  D_{\mathrm{KL}}(  P_{\text{aux}, l}(y_u| \textbf{x}_{1:t}  ,  y_{1:(u-1)}  )    \|  P (y_u| \textbf{x}_{1:t}  ,  y_{1:(u-1)}  )  \big]
%  ( D_{\mathrm{KL}}( P(y_u| \textbf{h}^{\text{enc}}_t, \textbf{h}^{\text{pre}}_u ) \| P_{\text{aux}}(y_u| \textbf{h}^{\text{enc, aux}}_t, \textbf{h}^{\text{pre}}_u ) \\
%  & \quad   +  D_{\mathrm{KL}}(  P_{\text{aux}}(y_u| \textbf{h}^{\text{enc, aux}}_t, \textbf{h}^{\text{pre}}_u )  \| P(y_u| \textbf{h}^{\text{enc}}_t, \textbf{h}^{\text{pred}}_u ) )
\end{split}  %  \qquad  \quad
\label{eq:kl}
\end{equation}
\noindent where the posteriors are given by Eq. \ref{eq:posterior_1} and \ref{eq:posterior_2}. 
Such KL divergence criterion can also guide the auxiliary branch with the supervision signals from primary branch, as a knowledge distillation procedure. As analyzed in \cite{li2020dynamic}, the gradients of multiple loss functions can be counteractive, and such KL loss penalizes the inconsistent gradients with respect to their shared parameters. Thus, the training objective can be: 
% improves the optimization consistency,
\begin{equation}
\begin{split}
   \mathcal{L} (\theta, \phi^{\Phi })  =  \mathcal{L}^{\text{\tiny{RNN-T}}}  (\theta) +
   \lambda_{\text{aux}}   \mathcal{L}^{\text{KL}} (\theta, \phi^{\Phi }) 
\end{split}
\label{eq:objective2}
\end{equation}
However, the direct application of RNN-T criterion to the auxiliary model can still be useful, since the auxiliary branch thus always contributes meaningful gradients before the primary model outputs are informative.  
Therefore, the overall training objective becomes:
\begin{equation}
\begin{split}
   \mathcal{L} (\theta, \phi^{\Phi }) & =  \mathcal{L}^{\text{\tiny{RNN-T}}}  (\theta) +
   \lambda_{\text{aux}} ( \mathcal{L}^{\text{\tiny{RNN-T}}}  ( \theta_{\text{shared}}^{\text{enc}}, \phi^{\Phi } ) + 
   \mathcal{L}^{\text{KL}} (\theta, \phi^{\Phi } ) )  
\end{split}
\label{eq:objective3} 
\end{equation}
\noindent Finally, after training, we discard the auxiliary branch and there is no additional computation overhead for ASR decoding.

% ------------------------------------------------------------------------------------------------------------------------------------------------
\subsection{Auxiliary context-dependent graphemic state prediction}
\label{ssec:ce_loss}

In an HMM-based phonetic hybrid ASR system, the triphone HMM states are tied via traditional decision tree clustering \cite{young1994tree}. 
Such a set of tied triphone HMM states – also referred to as context-dependent phonetic states or senones \cite{dahl2012context} – are used as the output units for the neural network based acoustic model. 
To further remove the need of a pronunciation lexicon, context-dependent graphemic hybrid models have been developed, and the tri-grapheme HMM states are tied instead. Accordingly, the neural network output units become tied tri-grapheme states, i.e. chenones \cite{le2019senones}, 
and the training criterion is cross entropy (CE) loss in conventional hybrid CE models. 

\begin{figure}[t]
\centering
\includegraphics[width=0.99\linewidth]{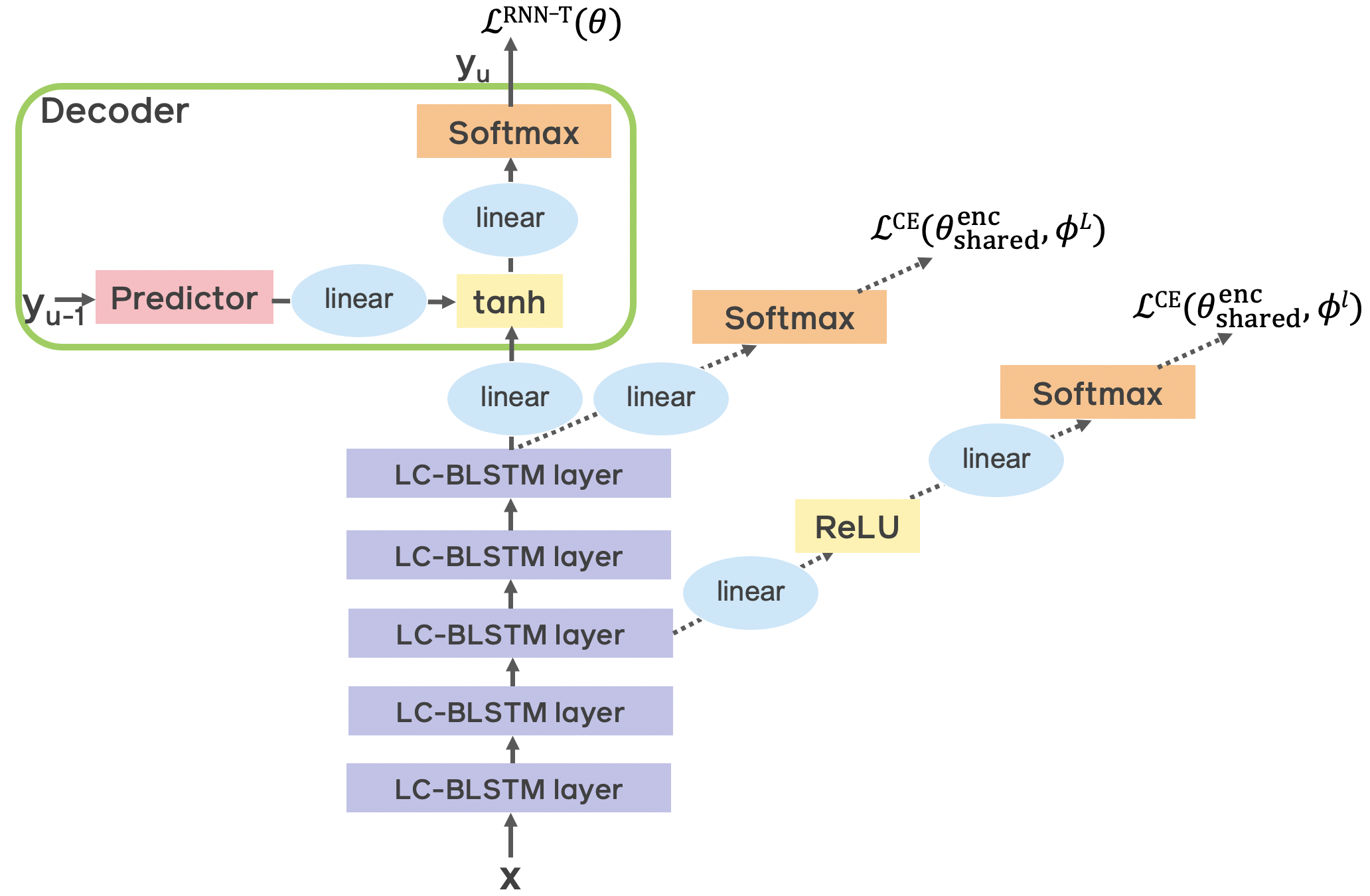}
\caption{\it Illustration of the proposed auxiliary context-dependent graphemic state  prediction task.}
\label{fig:ce}
\end{figure}

While RNN-T uses context-independent graphemes or wordpieces as output units, 
adding the chenone prediction supervision to encoder layers can transfer complementary tri-grapheme knowledge, encouraging diverse and discriminative encoder representations. 
% Since the cross entropy (CE) loss computation requires less memory than the auxiliary RNN-T loss during training, 
Then we can apply such CE criterion to multiple encoder layers.  
% In conjunction with the gradients from top-most RNN-T supervision  signals    notably  exploit  
Similarly, given an $L$-layer encoder,  denote $\Phi$ as a set of encoder layer indices that are connected to chenone prediction, and $\Phi \subseteq  \{ 1 \ldots L \}$. 
Denote $I$ as a binary indicator function, and  $1 \leqslant l \leqslant L$. 
% as 
% \begin{equation}
%\quad  = \left\{ 
%    I(l) = \left\{  \begin{array}{ll}  1,  \quad    l \in    \Phi    \\   % \hspace{1.82cm}  
%                  0,  \quad    l \notin \Phi     \\   \end{array}    \right. \end{equation}
% \noindent where  $1 \leqslant l \leqslant L$. 
As in Figure \ref{fig:ce}, if $I(l) = 1$, 
$\textbf{h}^{\text{enc}, l}$ goes through a one-hidden-layer multi-layer perceptron (MLP)\footnote{
Note that we use a linear layer rather than a MLP for the topmost/$L$th layer, since the top encoder layer has been designed for final label prediction.
}, parameterized by $\phi ^l$,  %   $\mathcal{W}$ 
and then a softmax function to provide a posterior distribution over chenone label set $\mathcal{S}$: 
\begin{equation}
       P( s_t |  \textbf{h}^{\text{enc}, l}_t  ) =   \text{Softmax}(\text{MLP}(\textbf{h}^{\text{enc}, l}_t ))
\end{equation}
\noindent where $s_t \in \mathcal{S}$, and the auxiliary CE loss is
% $\textbf{s} = (s_1 \ldots s_T)$ is a chenone sequence and 
\begin{equation}
\mathcal{L}^{\text{\tiny{CE}}} (\theta_{\text{shared}}^{\text{enc}}, \phi^{  \Phi }) = 
  -    \frac{ 1 }{ T } \sum_{t = 1}^{T}   \sum_{l = 1}^{L}   I(l)  \log  P(s_t | \textbf{h}^{\text{enc}, l}_t ) 
\end{equation}
\noindent 
% $o_k$ are the output and the target for label $k$, and  $K$ is the number of unique chenone labels. 
The overall training objective is: 
\begin{equation}
\mathcal{L}(\theta, \phi^{\Phi } ) =  
   \mathcal{L}^{\text{\tiny{RNN-T}}} (\theta) + \lambda_{\text{ce}} \mathcal{L}^{\text{\tiny{CE}}} ( \theta_{\text{shared}}^{\text{enc}}, \phi^{\Phi } )  
\label{eq:objective4}    % \lambda_{\text{\tiny{ce}}} 
\end{equation}

\noindent where $\lambda_{ce}$ is a tunable weighting parameter.

% ------------------------------------------------------------------------------------------------------------------------------------------------
% ------------------------------------------------------------------------------------------------------------------------------------------------

\section{Experiments}
\label{sec:exp}

\subsection{Experimental setup}
\label{ssec:setup}

\subsubsection{Data}
\label{sssec:data}

% ------------------------------------------------------------------------------------------------------------------------------------------------
%\setlength{\tabcolsep}{0.082cm}
%\renewcommand{\arraystretch}{1.0}
\begin{table}[t]
\caption{\label{tab:data} {\it  The amounts of audio data in hours.}}
\centering 
\begin{tabular}{   c  c  c c c c  c }
\hline \hline
        Language               &       Train  & Valid  &   \multicolumn{2}{   c  }{Test }  \\ 
                                &             &        &   clean         &     noisy     \\
\hline \hline
\multirow{1}{*}{ }    Romanian         & 161       &      5.2     &    5.1     &   10.2    \\
 			         Turkish          & 3.1K       &     13.6      &     21.2  &  23.4  \\    
                    German            & 3.2K      &     13.8    &     24.5     &  24.0 \\  
\hline \hline
\end{tabular}
\end{table}
% ------------------------------------------------------------------------------------------------------------------------------------------------

We first evaluate our proposed approaches on our in-house Romanian, Turkish and German video datasets, which are sampled from public social media videos and de-identified before transcription.  
These videos contain a diverse range of acoustic conditions, speakers, accents and topics. % making ASR challenging.   
The test sets for each language are composed of \emph{clean} and \emph{noisy} categories, with \emph{noisy} category being more acoustically challenging than \emph{clean}. The dataset sizes are shown in Table \ref{tab:data}.
Moreover,  we also perform evaluations on the public LibriSpeech dataset \cite{panayotov2015librispeech}. 

Input acoustic features are 80-dimensional log-mel filterbank coefficients with 25 ms window size, and we apply mean and variance normalization.
%  and a 10 ms time step between two windows    for each language
We apply the frequency and time masking as in the policy LD from SpecAugment \cite{park2019specaugment}, with $p=0.2$ on video datasets and $p=1.0$ on LibriSpeech. 
We perform speed perturbation \cite{ko2015audio} of the training data, and produce three versions of each audio with speed factors $0.9$, $1.0$ and $1.1$. The training data size is thus tripled. 
For the low-resource Romanian, we further apply another 2-fold data augmentation based on additive noise as in \cite{liu2020multilingual}, and the training data size is thus 6 times the size of original train set. 

\subsubsection{System implementation details}
\label{sssec:system}

For each video language, RNN-T output labels consist of a blank label and 255 wordpieces generated by the unigram language model algorithm from SentencePiece toolkit \cite{kudo2018sentencepiece}.
To provide chenone labels (Section \ref{ssec:ce_loss}), forced alignments are generated via a graphemic hybrid model \cite{le2019senones} for each language, and the number of unique chenone labels range from 7104 to 9272. 

% ro 9272, tr 7104, de 8334 

% For each utterance we compute its log mel spectrogram with $\nu$ dimension   and  $\tau$  time steps:  
% Frequency masking is applied $m_F$  times, and each time the frequency bands $\lbrack f_0$, $f_0+ f)$ are masked, where $f$ is sampled from $\lbrack 0, F\rbrack$ and  $f_0$ is sampled from $\lbrack 0, \nu - f)$.     
%  Time masking is optionally applied $m_T$ times, and each time the time steps $\lbrack t_0$, $t_0+ t)$ are masked,  where  $t$ is sampled from $\lbrack 0, T\rbrack$ and   $t_0$   is sampled from $\lbrack 0, \tau - t)$.   
% emulate mean shifts in spectrum  
%   anonymized.  We categorized them into sets    \texttt{clean} and \texttt{noisy} 

For video datasets, we build each RNN-T encoder based on latency-controlled bidirectional long short-term memory (LC-BLSTM) network  \cite{zhang2016highway}. 
Each encoder is a 5-layer LC-BLSTM network with 800 hidden units in each layer and direction, and dropout 0.3. 
Two subsampling layers with stride 2 are applied after first and second LC-BLSTM layer. 
The prediction network is a 2-layer LSTM of 160 hidden units for Romanian, and 512 units for Turkish and German, with dropout $0.3$. 
Each joint network has 1024 hidden units, and a softmax layer of 256 units for blank and wordpieces. 
For all neural network implementation, we use an in-house extension of PyTorch-based \emph{fairseq} \cite{ott2019fairseq} toolkit. 
All experiments use multi-GPU and mixed precision training supported in \emph{fairseq}, Adam optimizer \cite{kingma2014adam}, and tri-stage \cite{park2019specaugment} learning rate schedule with peak learning rate $4e^{-4}$.

For LibriSpeech, we experiment with two VGG transformer encoders of 24 and 36 layers as in \cite{wang2019transformer}, except that we use three VGG blocks with stride 2 in the first two blocks and 1 in the third block. 
Each transformer layer has an embedding dimension 512 and attention heads 8;  
feed-forward network (FFN) size is 2048 for 24-layer transformer, and 3072 for the 36-layer.
Wordpiece size is 1000 for the 24-layer, and 2048 for the 36-layer, 
resulting in total model parameters of 83.3M and 160.3M respectively. 

% efficient half precision floating point (FP16)     with betas (0.9, 0.999)  and epsilon $1e^{-8}$
% ------------------------------------------------------------------------------------------------------------------------------------------------
% ------------------------------------------------------------------------------------------------------------------------------------------------
\setlength{\tabcolsep}{0.14cm}
%\renewcommand{\arraystretch}{1.0}
%\begin{table*}[t]
\begin{table}[t]
\caption{\label{tab:results_ro_aux} 
{\it  WER results on Romanian dataset.  $\lambda_{\text{aux}}$ is used in Eq. \ref{eq:objective1}, \ref{eq:objective2} and \ref{eq:objective3}.
 ``aux" and ``kl" loss denote the auxiliary RNN-T (Section \ref{ssec:aux_loss}) and KL divergence criterion (Section \ref{ssec:kl_loss}) respectively.  
``crosslingual pretrain" denotes the encoder pretrained from a high-resource Spanish RNN-T. 
WERR (\%) is the unweighted average of the respective relative WER reductions on clean and noisy test sets.    } }
\centerline{ 
% \begin{tabular}{  p{3.0cm}  | c | p{0.8cm} |  p{0.8cm}  p{0.8cm} | c  }     In Section \ref{ssec:aux_loss} 
\begin{tabular}{  p{3.2cm}  | c   c|  c c  c  }
\hline \hline
         Model                      & $\lambda_{\text{aux}}$  & valid    & clean   & noisy  &  \footnotesize{WERR}  \\ 
\hline 
  baseline                          &  --    &  24.0    &  20.5   &  22.0  &  --     \\  \cdashline{1-6}[1.0pt/0.5pt]
                                    &  0.1   &  23.2    &         &        &         \\    
    \quad \texttt{+} aux loss       & \textbf{0.3}  & \textbf{22.8} &  19.6   &  21.0  &  4.5\%   \\
                                    &  0.6   &  23.1    &         &        &         \\    \cdashline{1-6}[1.0pt/0.5pt]
                                    &  0.3   &  22.9    &         &        &         \\ 
    \quad \texttt{+} kl loss        &  \textbf{0.6}  & \textbf{22.6} &  19.3   &  20.6  &  6.1\%   \\
                                    &  0.9   &  22.7    &         &        &          \\   \cdashline{1-6}[1.0pt/0.5pt] 
    \quad \texttt{+} aux \texttt{+} kl loss                &  0.3   &  22.5    &  19.1   &  20.8  &  6.1\%   \\ \cdashline{1-6}[4.0pt/0.5pt]
    \quad \texttt{+} crosslingual pretrain                           &  --    &  19.4    &  15.9   &  17.6  &  21.2\%   \\
    \qquad \texttt{+}  aux \texttt{+} kl loss   &  0.3   &  18.9    &  15.7   &  17.2  &  22.6\%  \\
\hline \hline
\end{tabular}}
\end{table}  

\setlength{\tabcolsep}{0.06cm}
%\renewcommand{\arraystretch}{1.0}
%\begin{table*}[t]
\begin{table}[t]
\caption{\label{tab:results_ro_ce} 
{\it  WER results on Romanian.  $\lambda_{\text{ce}}$ is used in  Eq. \ref{eq:objective4}.
``ce pretrain" denotes encoder pretraining from graphemic hybrid CE model.  
``ce loss" denotes auxiliary chenone prediction objective function (Section \ref{ssec:ce_loss}).    
``mid" denotes connecting CE loss to the 3rd (middle) encoder layer, and ``top" denotes connecting CE loss to the 5th (topmost) encoder layer.}}
\centerline{ 
% \begin{tabular}{  p{3.0cm}  | c | p{0.8cm} |  p{0.8cm}  p{0.8cm} | c  }
\begin{tabular}{  p{4.2cm}  | c   c|  c c  c  }
\hline \hline
         Model                        & $\lambda_{\text{ce}}$   & valid   & clean   & noisy  & \footnotesize{WERR}  \\ 
\hline 
  baseline                            &  --   &  24.0    &  20.5   &  22.0  &  --     \\   
   \quad \texttt{+} ce pretrain       &  --   &  22.8    &  19.3   &  20.9  &  5.4\%  \\     \cdashline{1-6}[1.0pt/0.5pt]
                                      &  0.3  &  23.2    &         &        &         \\   
   \quad \texttt{+}  ce loss, top     &  \textbf{0.6}  &  \textbf{22.9}   &  19.8   &  21.2  &  3.5\%  \\ 
                                      &  0.9  &  23.1    &         &       &          \\    \cdashline{1-6}[1.0pt/0.5pt]
                                      &  0.3  &  22.3    &         &       &          \\  
   \quad \texttt{+}  ce loss, mid     &  \textbf{0.6}   &   \textbf{22.0}  &    18.5     &  20.3     &  8.7\%        \\
                                      &    0.9          &  22.0            &            &   &     \\  \cdashline{1-6}[1.0pt/0.5pt]
   \quad \texttt{+} ce pretrain, ce loss, mid       &  0.6  &  21.4    &  17.9   & 19.6  &  11.8\%  \\  \cdashline{1-6}[1.0pt/0.5pt]
   \quad \texttt{+} ce pretrain, ce loss, mid, top  &  0.6  &  21.2    &  17.8   & 19.5  &  12.3\%  \\  \cdashline{1-6}[1.0pt/0.5pt]
\hline \hline
\end{tabular}}
\end{table}
% ------------------------------------------------------------------------------------------------------------------------------------------------

% ------------------------------------------------------------------------------------------------------------------------------------------------
\setlength{\tabcolsep}{0.06cm}
\begin{table}[t]
\caption{\label{tab:results_tr_de} {\it WER results on Turkish and German, with $\lambda_{\text{aux}}=0.3$ and $\lambda_{\text{ce}}=0.6$. }}
\centerline{ 
\begin{tabular}{  p{3.1cm} |c c c |c c c }
\hline \hline
             &     \multicolumn{3}{ | c  }{Turkish }   &     \multicolumn{3}{ |c  }{German }          \\ 
   Model     &  clean   &   noisy     &  \footnotesize{WERR} &  clean   & noisy     &  \footnotesize{WERR}  \\
\hline 
 baseline                         &   17.1   &  18.9 &   --        &   11.6    &  13.0 & --      \\  % \cdashline{1-7}[1.0pt/0.5pt] 
\hline 
  \quad \texttt{+} aux loss       &   16.8   &  18.8 &  1.1\%      &   11.3    &  12.6 & 2.8\%   \\   
  \quad \texttt{+} kl loss        &   16.7   &  18.8 &  1.4\%      &   11.5    &  12.8 & 1.2\%   \\ 
  \quad \texttt{+} aux \texttt{+} kl loss    &   16.4   &  18.5 & \textbf{3.1\%}  &  11.3    &  12.6 & \textbf{2.8\%}   \\ \cdashline{1-7}[1.0pt/0.5pt] 
  \quad \texttt{+} crosslingual pretrain       &   16.6   &  18.6 &  2.3\%      &   11.4    &  12.8 & 1.6\%   \\ 
  \qquad \texttt{+} aux \texttt{+} kl loss  &  16.1 &  18.1 & \textbf{5.0\%}  &  11.3 & 12.4 & \textbf{3.6\%}\\ \cdashline{1-7}[1.0pt/0.5pt] 
\hline 
\quad \texttt{+} ce pretrain                 & 16.8  & 18.9 &  0.9\%          & 11.5  & 12.8   & 1.2\%  \\ 
  \qquad \texttt{+} ce loss, mid            & 16.5  & 18.4 &  3.1\%          & 11.3  &  12.5  & 3.2\%  \\ 
  \qquad \texttt{+} ce loss, mid, top       & 16.3  & 18.2 & \textbf{4.2\%}  &  11.2 & 12.3   & \textbf{4.4\%}  \\ \cdashline{1-7}[1.0pt/0.5pt]  
\hline \hline
\end{tabular}}
\end{table}
% ------------------------------------------------------------------------------------------------------------------------------------------------

\subsection{Auxiliary RNN-T modeling results on video datasets}
\label{ssec:aux_results}

We first perform experimental evaluations on the low-resource language Romanian, and obtain the optimal $\lambda_{\text{aux}}$ in Eq. \ref{eq:objective1}, \ref{eq:objective2} and \ref{eq:objective3}. 
% (Section \ref{ssec:aux})
ASR word error rate (WER) results are shown in Table \ref{tab:results_ro_aux}. 
For both \emph{clean} and \emph{noisy} test sets, we first compute the relative WER reduction (WERR) over respective baseline as a percentage, and then take the unweighted average of two percentages, which we refer to as an average WERR. 

As shown in Table \ref{tab:results_ro_aux}, for auxiliary RNN-T loss (Eq. \ref{eq:objective1}), we vary $\lambda_{\text{aux}}$ over $\{$0.1, 0.3, 0.6$\}$, and observe $0.3$ gives the lowest WER on valid set. So we proceed with $\lambda_{\text{aux}}=0.3$ to decode the clean and noisy test sets, and see an average WERR $4.5\%$. Similarly for the symmetric KL divergence loss (Eq. \ref{eq:objective2}), we vary $\lambda_{\text{aux}}$ over $\{$0.3, 0.6, 0.9$\}$; we find $\lambda_{\text{aux}} = 0.6$ works best and provides an average WERR $6.1\%$. 
When combining the two objectives with $\lambda_{\text{aux}}=0.3$ (Eq. \ref{eq:objective3}), we find it also gives an average WERR $6.1\%$, which is better than using auxiliary RNN-T loss on its own.

For the low-resource scenario, one approach to address the lack of resources are to make use of data from high-resource languages. We thus perform crosslingual pretraining experiments with a Spanish RNN-T model trained on 7K hours. We use the Spanish encoder as the pretrained encoder for Romanian, and proceed with RNN-T training as before, which provides substantial improvements as in Table \ref{tab:results_ro_aux}. 
While on top of crosslingual pretraining, adding auxiliary RNN-T and KL divergence loss provides moderate gain. 

We use the optimal $\lambda_{\text{aux}}$ found in each condition and evaluate the performance on Turkish and German.
As shown in Table \ref{tab:results_tr_de}, the proposed combination of auxiliary RNN-T and KL divergence loss provides consistent improvements, which is also better than using each individually. 
We use the same Spanish encoder for crosslingual pretraining, and the improvements are much less due to the increased training data size. 
Along with the proposed auxiliary RNN-T modeling, they combine to produce noticeable gains. 

% ------------------------------------------------------------------------------------------------------------------------------------------------
\subsection{Auxiliary chenone prediction results on video datasets}
\label{ssec:ce_results}

Since we build graphemic hybrid systems to provide chenone labels, we can additionally use the hybrid model as pretrained encoder for RNN-T. 
As shown in \cite{rao2017exploring, hu2020exploring}, pretraining RNN-T encoder with connectionist temporal classification (CTC) or hybrid CE criterion can improve performance, and we also find CE pretraining produces an average 5.4\% WERR on the low-resource Romanian as in Table \ref{tab:results_ro_ce}.

For the medium-resource Turkish and German (i.e. training data size of {\fontfamily{pcr}\selectfont\texttildelow}3K hours), we initially find pretraining with hybrid CE model can provide 2 - 4\% improvements with a relatively small training mini-batch size. However, after optimizing the memory cost by mixed precision training and function merging \cite{li2019improving}, RNN-T training can enable larger mini-batch size, and we only observe minor improvements 0.9 - 1.2\% in Table \ref{tab:results_tr_de}. 

Then we experiment with $\lambda_{\text{ce}}$ (Eq. \ref{eq:objective4}) on Romanian. 
Given each 5-layer LC-BLSTM encoder, we also examine connecting chenone prediction to 3rd (middle) layer or 5th (topmost) layer. 
As in Table \ref{tab:results_ro_ce}, $\lambda_{\text{ce}}=0.6$ works best in each case. While attaching chenone prediction to middle layer performs better than top layer, they combine to provide further improvements on top of CE pretraining.

We continue to evaluate the Turkish and German performance with $\lambda_{\text{ce}}=0.6$. As in Table \ref{tab:results_tr_de}, 
training on both middle and top layers for auxiliary chenone prediction outperforms training on each alone, and produces noticeable improvements of 4.2 - 4.4\% when combined with CE pretraining. 

% we observe that for medium-resource languages , CE pretraining only gives minor improvement of 0.9 - 1.2\%. 
% We use the model after hybrid CE training, and refer to it as  in Table   effectively produces $5.4\%$ WERR.
% the forced alignments are generated via each graphemic hybrid model   noticeable 
%  Also, connecting chenone prediction to both middle and top encoder layers perform better than each alone    Specifically 
% ------------------------------------------------------------------------------------------------------------------------------------------------
% ------------------------------------------------------------------------------------------------------------------------------------------------
\setlength{\tabcolsep}{0.14cm}
%\renewcommand{\arraystretch}{1.0}
%\begin{table*}[t]
\begin{table}[t]
\caption{\label{tab:aux_loss_libri} 
{\it  WER results on LibriSpeech, with 24-layer transformer encoder and 83M total model parameters.    } }
\centerline{ 
% \begin{tabular}{  p{3.0cm}  | c | p{0.8cm} |  p{0.8cm}  p{0.8cm} | c  }     In Section \ref{ssec:aux_loss} 
\begin{tabular}{  p{2.8cm}  | c   c|  c c  c  }
\hline \hline
         Model                               & test-clean &  \footnotesize{WERR}  & test-other    &  \footnotesize{WERR}  \\ 
\hline 
  baseline                                      &  2.77   &  --         &  6.60   &  --    \\  \cdashline{1-6}[1.0pt/0.5pt]
    \quad \texttt{+} aux \texttt{+} kl loss     &  2.48   &  10.6\%     &  5.62   & 14.8\%   \\    \cdashline{1-6}[1.0pt/0.5pt]
    \quad \texttt{+} ce loss                    &  2.42   & 12.6\%      & 5.75    & 12.9\%   \\  \cdashline{1-6}[1.0pt/0.5pt] 
    \quad \texttt{+} aux \texttt{+} kl \texttt{+} ce loss   & 2.31  & 16.5\%  & 5.26   & 20.3\% \\ \cdashline{1-6}[4.0pt/0.5pt]
\hline \hline
\end{tabular}}
\end{table}  
% ------------------------------------------------------------------------------------------------------------------------------------------------
% ------------------------------------------------------------------------------------------------------------------------------------------------
\setlength{\tabcolsep}{0.07cm}
\begin{table}[h]
\caption{\label{tab:results_libri} 
{\it  Comparison of our models (with 36-layer transformer encoder and 160M total model parameters) with recently published best results on LibriSpeech.    } }
\centerline{ 
% \begin{tabular}{  p{3.0cm}  | c | p{0.8cm} |  p{0.8cm}  p{0.8cm} | c  }     In Section \ref{ssec:aux_loss} 
\begin{tabular}{  p{2.9cm}   c  c  c c  c  }
\hline \hline
 Model                               & \multicolumn{2}{ c  }{w/o LM }   &  \multicolumn{2}{ c  }{w/ LM }   \\   \cdashline{2-6}[1.0pt/0.5pt]
                                     & test-clean &   test-other   & test-clean  &  test-other    \\ 
\hline 
 \textbf{LAS}                                &       &        &        &      \\ 
   LSTM  \cite{park2020specaugment}          &  2.6  &  6.0   &  2.2   & 5.2  \\  \cdashline{1-6}[4.0pt/0.5pt]
 \textbf{Hybrid}                             &       &        &        &      \\ 
   Transformer \cite{wang2019transformer}    &  2.6  &  5.6   &  2.3   & 4.9  \\  \cdashline{1-6}[1.0pt/0.5pt]
 \textbf{CTC}                                &       &        &        &      \\ 
   Transformer \cite{zhang2020fast}          &  2.3  &  4.8   &  2.1   & 4.2  \\  \cdashline{1-6}[4.0pt/0.5pt]
 \textbf{Sequence Transducer}                &       &        &        &      \\ 
   Transformer \cite{zhang2020transformer}   &  2.4  &  5.6   &  2.0   & 4.6  \\ 
   % ContextNet \cite{han2020contextnet}       &  2.1  &  4.6   &  1.9   & 4.1  \\
   Conformer   \cite{gulati2020conformer}    &  2.1  &  4.3   &  1.9   & 3.9  \\ 
   Transformer (\textbf{Ours})               &  \textbf{2.2}     & \textbf{4.7}  &   \textbf{2.0}      & \textbf{4.2}  \\ 
\hline \hline
\end{tabular}}
\end{table}  
% ------------------------------------------------------------------------------------------------------------------------------------------------
% ------------------------------------------------------------------------------------------------------------------------------------------------
\subsection{Results on LibriSpeech with transformer encoders}

% Follow similar setup in \cite{zhang2020fast}, we use a 24-layer Transformer as encoder when performing experiments on Librispeech. 
% Since the encoder is much deeper than  studied above,  and apply KL divergence against the main joiner output using the 24th layer encoding
While we use streamable 5-layer LC-BLSTM encoders on video datasets above, we experiment with 24/36-layer transformer encoders instead on LibriSpeech. Given the much larger encoder depth, when evaluating the auxiliary RNN-T and KL divergence,  we find it more effective to apply the loss at multiple layers. Thus for the 24-layer transformer, we apply it to the \nth{6}, \nth{12} and \nth{18} encoder layers. 
As in Table \ref{tab:aux_loss_libri}, it provides about 11\% and 15\% WERR on each test set. 
When evaluating the auxiliary CE loss, we apply it at the middle (\nth{12}) and top (\nth{24}) layer again, which also produces substantial relative gains about 13\%. 

Additionally, we also attempt to apply both auxiliary tasks simultaneously, i.e., auxiliary RNN-T and KL divergence loss at \nth{6} and \nth{18} layers, and CE loss at \nth{12} and \nth{24} layers. 
In all cases, we use $\lambda_{\text{aux}} = 0.3$ and $\lambda_{\text{ce}}=0.6$ found above (Section \ref{ssec:aux_results} and \ref{ssec:ce_results}). 
As in Table \ref{tab:aux_loss_libri}, both auxiliary tasks combine to produce significant and complementary improvements. These performance gains are much larger than those on the video datasets with 5-layer LC-BLSTM encoder. We conjecture that transformer networks of increased depth suffer more from the encoder undertraining and gradient vanishing problem at lower layers (as discussed in Section \ref{ssec:aux}), and auxiliary tasks play more effective roles in addressing it.  

We proceed to increase transformer encoder depth from 24 to 36 layers, FFN size from 2048 to 3072, and wordpiece size from 1000 to 2048. 
We observe that without the auxiliary tasks, neither 24-layer transformer of FFN size 3072 nor 36-layer transformer of FFN 2048 is able to converge. Instead both can converge while using either of the two auxiliary tasks. 
Finally, the 36-layer transformer of FFN 3072 - which uses auxiliary RNN-T and KL divergence loss at \nth{9} and \nth{27} layers, and CE loss at \nth{18} and \nth{36} layers -  produces our best results in Table \ref{tab:results_libri}.
Auxiliary tasks thus provide an opening for learning deep encoder network, and the increased depth is central to accuracy gains.

We further perform first-pass shallow fusion \cite{kannan2018analysis} with an external language model (LM). 
We use a 4-layer LSTM LM with 4096 hidden units, and LM training data consists of LibriSpeech transcripts and text-only corpus (800M word tokens), tokenized with the 2048 wordpiece model. As in Table \ref{tab:results_libri}, we thus achieve competitive results compared to the prior top-performing models. 

% Lastly, we trained a larger model with 36-layer Transformer as encoder and combined with shallow fusion, we achieved best WER on test-clean and test-other as shown in Table \ref{tab:results_libri}.
% we observe significant improvements by using auxiliary RNN-T and CE loss alone, each achieved greater than 10\% WERR. By using both auxiliary RNN-T and chenone prediction criteria, we obtained over 20\% WERR on test-others.
% we use all 4 encoder intermediate layer output mentioned above, where the 6th and 18th layer output were used to compute transducer and kl aux loss while 12th and 24th output were used to compute CE aux loss. 
% We followed similar loss weighting scheme as LC-BLSTM, where weight of transducer and kl aux loss is 0.3 and CE aux loss is 0.6.
% ------------------------------------------------------------------------------------------------------------------------------------------------
% ------------------------------------------------------------------------------------------------------------------------------------------------
\section{Related Work}
\label{sec:related}

Attaching auxiliary objective functions to intermediate layers has been explored in various prior works.  
For improving image recognition, multiple auxiliary classifiers with squared hinge losses were used in \cite{lee2015deeply}, and CE objective functions used in \cite{szegedy2015going, szegedy2016rethinking}, while later \cite{szegedy2016rethinking} only reported limited performance gain. 
\cite{li2020dynamic} made further progress by showing that, the gradients of multiple loss functions with respect to their shared parameters can counteract each other, and minimizing the symmetric KL divergence between the multiple classifier outputs can penalize such inconsistent gradients and provide more performance gains. 

Similarly, for improving hybrid ASR models trained with CE criterion, \cite{lu2019self} connected an intermediate layer directly with a linear projection layer  to compute the logits over senones, and used an asymmetric KL divergence loss between the primary model output (i.e. senone posterior) and the auxiliary classifier output. 
While in our work, we found connecting a nonlinear MLP - rather than a single linear layer - to the intermediate layer is more effective, which disentangled  low-level feature extraction from  final wordpiece prediction. 
Also for improving ASR, \cite{tjandra2020deja} applied CTC or CE objective functions to multiple encoder layers, although without the cross-layer KL divergence loss. 
While CTC or hybrid senone/chenone models can directly produce posteriors over output labels, RNN-T requires a decoder to compute the output (wordpiece) posterior. Thus, in applying the auxiliary RNN-T or KL divergence loss, we specifically share the RNN-T decoder during the forward pass while keeping it intact from the backward pass (as discussed in Section \ref{ssec:aux_loss}).

% Because the auxiliary loss is to address the encoder underfitting issue, decoder is not explicitly learned to fit the auxiliary objective.
% to address the potential RNN-T encoder undertraining issue. 

Note that compared to CTC or hybrid models, attention-based seq2seq model is more similar to RNN-T, since both have a neural encoder and decoder. 
And for attention-based seq2seq model, using auxiliary senone labels has shown improved WERs in \cite{toshniwal2017multitask, moriya2018multi}, while recent work \cite{inaguma2020minimum} showed contrary observations.

% As discussed in  Section \ref{ssec:kl_loss}     focused on model pretraining
% ------------------------------------------------------------------------------------------------------------------------------------------------
\section{Conclusions}

In this work, we propose the use of auxiliary tasks in improving RNN-T based ASR. 
We first benchmark the streamable LC-BLSTM encoder based performance on video datasets. 
Applying either auxiliary RNN-T or symmetric KL divergence objective function to intermediate encoder layers has been shown to improve ASR performance, and combining both is more effective than each on its own. 
Performing auxiliary chenone prediction also provides noticeable complementary gains on top of hybrid CE pretraining.  

Next, we demonstrate the efficacy of both auxiliary tasks in improving the transformer encoder based sequence transducer results on LibriSpeech. 
Both auxiliary tasks provide substantial and complementary gains, and we find that, critical to the convergence of learning deep transformer encoders is the application of auxiliary objective functions to multiple encoder layers. 
Lastly, to participate in the LibriSpeech benchmark challenge, we develop a 36-layer transformer encoder via both auxiliary tasks, which achieves a WER of 2.0\% on test-clean, 4.2\% on test-other.

%  compared to CE pretraining. 
% crosslingual and hybrid CE pretraining  

%\vfill
%\pagebreak

%\section{REFERENCES}
%\label{sec:ref}

\bibliographystyle{IEEEbib}
\bibliography{thesis,refs}

\end{document}